\documentclass[letterpaper]{article}
\usepackage{aaai}
\usepackage{times}
\usepackage{helvet}
\usepackage{courier}
\usepackage{multirow}
\usepackage{amsmath}
\usepackage{amssymb}

\usepackage{rotating}
\usepackage[table]{xcolor}
\usepackage{array,hhline}
\makeatletter
\newcommand*{\ccol}[1]{%
  \ifdim#1pt<.5pt\relax\else\color{white}\fi
  \edef\x{\noexpand\cellcolor[gray]{\strip@pt\dimexpr1pt-#1pt}}\x
  #1%
}
\newlength{\cellwidth}
\makeatother

\nocopyright

\begin{document}
%
\title{Action Recognition in the Frequency Domain\thanks{This work was partially supported by the DARPA MindÕs Eye program. The U.S. Government is authorized to reproduce and distribute reprints for Governmental purposes notwithstanding any copyright annotation thereon. The views and conclusions contained herein are those of the authors and should not be interpreted as necessarily representing the official policies or endorsements, either expressed or implied, of DARPA or the U.S. Government.}}
\author{Anh Tran$^\dagger$,  Jinyan Guan$^\dagger$,  Thanima Pilantanakitti$^\ddagger$,  Paul Cohen$^\ddagger$\\
University of Arizona\\
$^\dagger$Department of Computer Science\\
$^\ddagger$School of Information: Science, Technology, and Arts\\
{\tt \{trananh, jguan1, tpilanta, prcohen\}@email.arizona.edu}
}
\maketitle
\begin{abstract}
\begin{quote}
In this paper, we describe a simple strategy for mitigating variability in temporal data series by shifting focus onto long-term, frequency domain features that are less susceptible to variability.   We apply this method to the human action recognition task and demonstrate how working in the frequency domain can yield good recognition features for commonly used optical flow and articulated pose features, which are highly sensitive to small differences in motion, viewpoint, dynamic backgrounds, occlusion and other sources of variability.  We show how these frequency-based features can be used in combination with a simple forest classifier to achieve good and robust results on the popular KTH Actions dataset.
\end{quote}
\end{abstract}

\section{Introduction}
\label{sec:introduction}

Human action recognition research has attracted considerable attention in recent years due to its practical applications in areas such as video surveillance, robotics, human-computer interaction, video indexing and retrieval, scene understanding and analysis, behavioral biometrics, biomechanics, and others.  

Typical recognition scenarios often include variations in motion, illumination and viewpoint, partial occlusions, variable execution rates and anthropometry of actors involved, changes in backgrounds, and so forth \cite{Aggarwal1999,Moeslund2001}.  These conditions pose great challenges for researchers and often induce much variability in the data.

In this paper, we present a strategy to reduce the effects on classifier performance of some of these kinds of variability. Rather than working with data in the temporal domain, it is sometimes better to extract features of the data in the frequency domain.  We apply this idea to two commonly used features in human action recognition research, optical flow and articulated pose, and show how their corresponding frequency-domain features can be effectively used for classification on the KTH Actions dataset from~\citeauthor{Schuldt2004} \shortcite{Schuldt2004}.  We adopt the efficient randomized forest-based approximate nearest-neighbor approach presented by \citeauthor{O'Hara2012} \shortcite{O'Hara2012} and apply it to our frequency-domain features to build \emph{frequency forest} classifiers.

Details of the frequency domain representation and classification method are described in Sections~\ref{sec:representation} and~\ref{sec:classification}, respectively.  Section~\ref{sec:related} presents related work, and Section~\ref{sec:experiments} describes our experimental procedures and results.  Section~\ref{sec:conclusion} concludes this paper.

\section{Related Work}
\label{sec:related}

There is a large body of literature on human action recognition research, detailing many innovative learning algorithms and novel representations for actions.  Several informative surveys are available, including \cite{Aggarwal1999,Moeslund2001,Gavrila1999,Wang2003,Turaga2008}.  Here, we will briefly review the more commonly used and recently developed representations and classifiers of actions.

A popular approach is to use localized space-time features extracted from video sequences.  \citeauthor{Schuldt2004} \shortcite{Schuldt2004} demonstrated that local measurements in terms of spatio-temporal interest points (STIP) can sufficiently represent the complex motion patterns of various actions.  Recent research also achieved success by representing \emph{tracklets} of actions as points on Grassmann manifolds, where each track is modeled as a 3-dimensional data cube with axes of width, height, and frame number \cite{O'Hara2012,Lui2010}.

Asides from low-level representations, there are work that focus on higher level representations of actions.  For instance, \citeauthor{Kerr2011} \shortcite{Kerr2011} described each action sequence as a collection of propositions that have truth values over some time intervals.  They showed that each action has a corresponding set of signature \emph{fluents} --- intervals during which propositions are true --- that can be learned, and that finite state machine models of actions can be constructed from these signatures.  Recently, \citeauthor{Sadanand2012} \shortcite{Sadanand2012} presented a new high-level representation of videos called Action Bank.  Inspired by the \emph{object bank} approach to image representation, an \emph{action bank} representation comprises the collected output of many pre-trained template-based action detectors that each produce a correlation volume.

Actions can also be modeled in terms of the optical flow of various regions of the image.  \citeauthor{Yacoob1998} \shortcite{Yacoob1998} represented actions using the optical flow of different body parts (e.g., torso, thighs, calfs, feet, and arms).  \citeauthor{Danafar2007} \shortcite{Danafar2007} segmented the body into more coarse regions (e.g., head-neck, body-hands, and legs) and represented actions using histograms of flow in these regions.

Innovations in pose estimation technology have also inspired representations centered on features extracted from articulated body parts.  In this area, some researchers like to work with articulated pose in 2D, as in \cite{Sheikh2005,Lv2007}; while others prefer to avoid the challenges of 2-dimensional image data by directly recording joints coordinates in 3D using the increasingly accessible RGBD cameras or other commercial motion capture systems, as in \cite{Campbell1995,Li2010,Wang2012,Sung2012,Xia2012}.

\section{Representation: Frequency Features}
\label{sec:representation}

Frequency features for action recognition are representations in the frequency domain that model the body motions associated with each action.  For example, the frequency features for a \emph{walk} action might capture the rhythmic swinging of the arms or the periodic movements of the legs, and not just the fact that a blob of energy shifted across the screen.  

\begin{figure}[!ht]
\centering
\fbox{\includegraphics[width=0.44\linewidth]{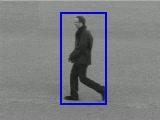}} ~
	\fbox{\includegraphics[width=0.44\linewidth]{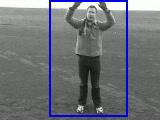}}
\caption{Example annotated bounding box localization for two different video frames.}
\label{fig:bbox}
\end{figure}

In this work, we assume that the body has been localized in the frame using some combinations of computer vision detection and tracking algorithms, or any other similar method.  This localization problem is itself an open challenge in computer vision.  However, it is not the focus of this paper, so the work reported here is done with hand-annotated videos in which bounding boxes have been drawn around the body.  (Some evidence suggests that feature-based methods require localization to perform well on datasets that have dynamic backgrounds \cite{Ryoo2010}, and that bounding-box style localization is easier to use than alternatives like silhouette or pose extraction.) Our bounding boxes were created using the VATIC annotation tool \cite{Vondrick2010}.\footnote{All annotations and relevant code are publicly available at: {\tt https://code.google.com/p/ua-gesture/}}  Examples of the boxes can be seen in Figure~\ref{fig:bbox}.  

The rest of this section explains how we represent actions in the frequency domain using optical flow and articulated pose features that are localized by bounding boxes.

\subsubsection{Optical Flow}

\begin{figure}[!ht]
\centering
\fbox{\includegraphics[width=0.44\linewidth]{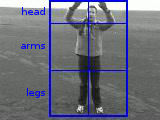}} ~
	\fbox{\includegraphics[width=0.33\linewidth]{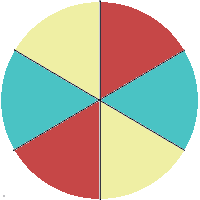}}
\caption{Illustration of optical flow partitioning for a bounding box.  The $5$ subregions representing different areas of the body are shown in the left image.  The right image shows the $6$ directional bins used to partition the flow within each body subregion.}
\label{fig:flow}
\end{figure}

We extract optical flow from images using the algorithm described in \citeauthor{Liu2009} \shortcite{Liu2009}.  Once extracted, we use the bounding box to localize the flow of the actor, and then further divide the box area into smaller regions.  Following a similar process to \citeauthor{Danafar2007} \shortcite{Danafar2007}, we split each bounding box into $5$ different subregions (see Figure~\ref{fig:flow}): head, left torso/arm, right torso/arm, left leg, and right leg.  The corresponding left and right regions are split evenly in the horizontal direction.  The head region occupies $1/5$ of the bounding box's height, while the torso/arms and legs regions are each $2/5$ of the height.

The flow within each subregion is binned into six different directions, as in Figure~\ref{fig:flow}.
Within each bin, the proportion of vectors that fall into the bin with respect to all vectors in the subregion, as well as the average magnitude of all flow vectors in the bin are computed.

These calculations yield {\em optical flow features} such as the proportion of flow vectors falling into the rightward bin in the head subregion, or the average magnitude of flow vectors falling into the up-leftward bin in the right leg subregion.  Optical flow features  are calculated for each frame, but we are interested in their time series across all frames, and particularly in transformations of these time series into the frequency domain by the Fourier transform.

\begin{figure*}[!ht]
\centering
\fbox{\includegraphics[width=0.47\linewidth]{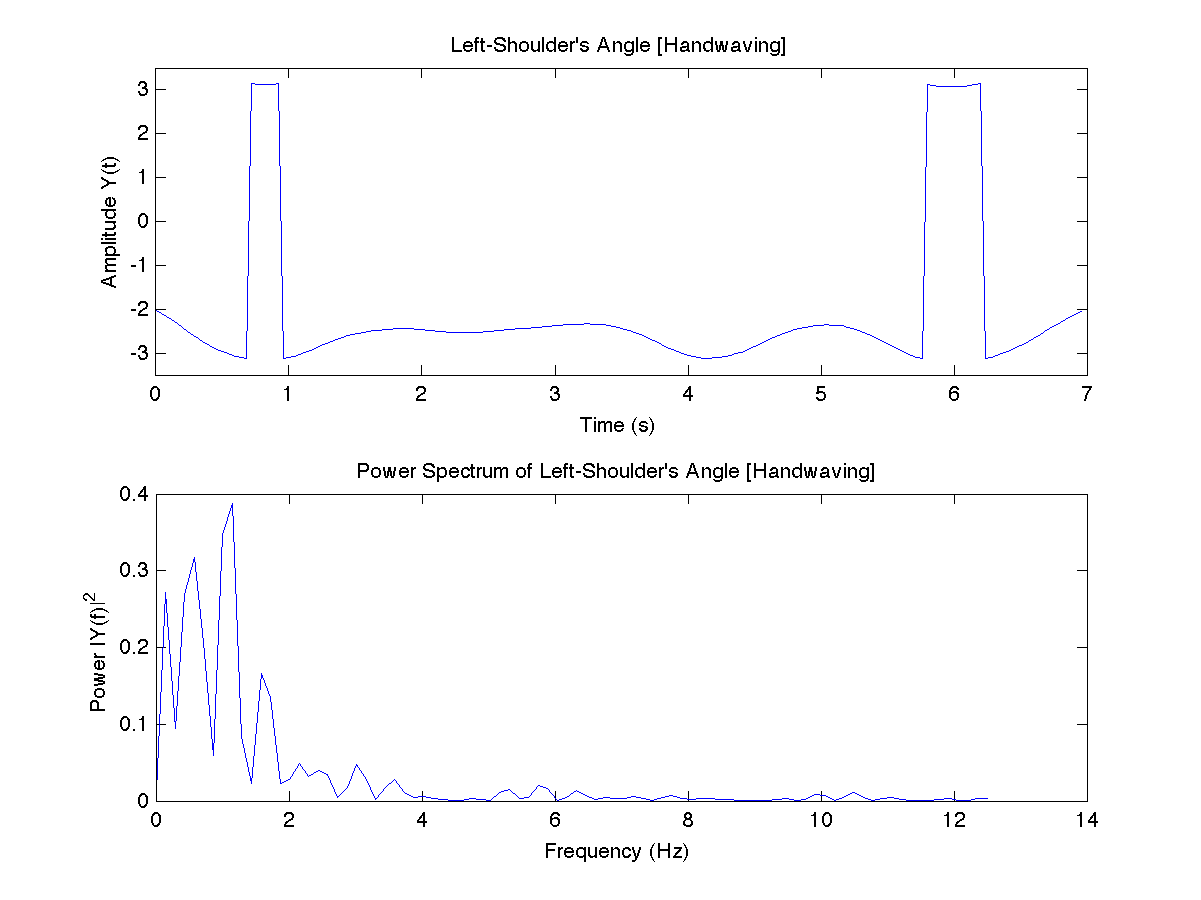}} ~
	\fbox{\includegraphics[width=0.47\linewidth]{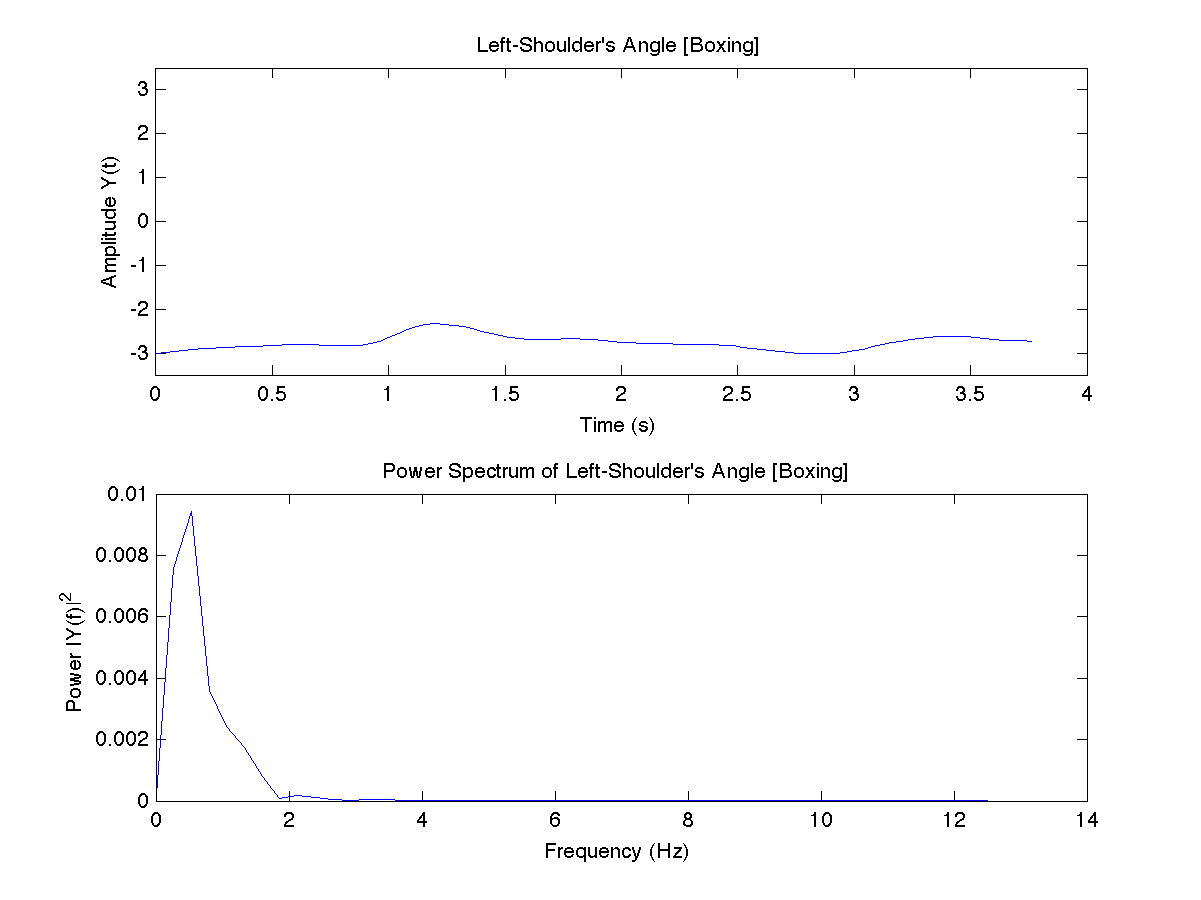}}
\caption{The smoothed left shoulder's angle for two different action sequences (\emph{handwaving} on the left and \emph{boxing} on the right) and the corresponding power spectra in the frequency domain.}
\label{fig:spectra}
\end{figure*}

After the Fourier transformation, we examine the power spectrum of each series and take the first $N$ components of the spectrum to be a frequency-domain feature. (In our experiments, $N=25$.) Figure~\ref{fig:spectra} shows some example time-series features and their corresponding power spectra.  For short video segments that yield fewer than $N$ components, we simply recycle the segment.  This method allows us to convert the time series of each optical flow feature into a corresponding frequency feature, and, in fact, we can calculate frequency features for any combination of optical flow features.  For example, among the 31 frequency features that we used for action recognition, we derived the right plus left flow proportions of each subregion, the upper-right plus lower-left proportions of each subregion, the lower-right plus upper-left proportions of each subregion, the average flow magnitudes in different directions of each subregion, and the average flow of the entire bounding box.

\subsubsection{Articulated Pose}

To extract articulated pose from images, we use the algorithm by \citeauthor{Yang2011} \shortcite{Yang2011} which returns all detected poses for each frame.  We then find the pose that best matches each bounding box (i.e., the pose with the highest detection score that fits in the area of the bounding box) to generate tracks of articulated poses.  We note that due to the low resolution of the data, some images needed to be enlarged to $2\times$ or $3\times$ magnification for the pose estimation algorithm to work properly.

\begin{figure}[!ht]
\centering
\fbox{\includegraphics[width=0.44\linewidth]{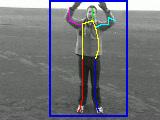}} ~
	\fbox{\includegraphics[width=0.44\linewidth]{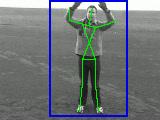}}
\caption{An example articulated pose output by \citeauthor{Yang2011} \shortcite{Yang2011} that best matches the bounding-box (left image) and the corresponding reformatted pose with fewer joints (right image).}
\label{fig:pose}
\end{figure}

We converted the $26$-joint pose format given by \citeauthor{Yang2011} \shortcite{Yang2011} to a simpler $15$-joint format that is more compatible with other pose datasets (see Figure~\ref{fig:pose}). The time series for each joint is smoothed and all poses are then standardized to be within a unit square.  This ensures that poses are all roughly equal in size, allowing some robustness to variability in the size of the figure in a frame. 

As with the optical flow features, we treat the different measurements, or relationships, between joint trajectories as temporal data series and convert them into corresponding frequency spectra using the Fourier transform.  We take the top $N = 25$ components of each associated power spectrum as a frequency feature.  For our experiments, we extracted a total of $15$ different frequency features from pose-based time series, such as the horizontal displacement between the left hand and left shoulder, vertical displacement between the left hand and the left shoulder, and various angles, like the angle of the left elbow.

\section{Classification: Frequency Forest}
\label{sec:classification}

We adopt the forest-based method presented by \citeauthor{O'Hara2012} \shortcite{O'Hara2012} for our classification model.  The data used, however, are frequency-domain features.  Based on the algorithm presented, trees are constructed by incrementally adding training samples to an initially empty root node.  Once a node becomes large enough and the splitting criteria are met, an element of the node is selected to be the pivot item and its distance to each item in the node is computed.  Items having a distance to the pivot that is less than or equal to some threshold are added to the left child node, and the rest to the right.  The now-empty node is marked as a splitting node and all subsequently added instances are forwarded to one of its children after being compared to the pivot.  The process recurses to form a tree.  In the end, all interior nodes are splitting nodes and all leaf nodes contain neighboring samples.

In our experiments, we use the Entropy splitting method, which dictates that a tree's node is only split when the distribution of distances between the items in the node falls below some empirically determined entropy threshold $t_e$ \cite{O'Hara2012}.  (For our system, $t_e = 1.79$.)  Furthermore, the Euclidean distance is used to measure the distance between two frequency features.

We build a frequency tree in the forest for each frequency feature.  Each tree is designed to return the top five nearest neighbors for each test instance.  However, only results from trees that have a distinct dominating label among the neighbors are considered for the final prediction.  That is, we only consider results from trees where at least three of the top five neighbors share the same label.  We believe this helps to prevent trees with weak correlations with the test instance from confusing the final prediction.  Once all trees in the forest have voted, we pool all valid results together and return the most popular label as the final prediction.

\section{Experiments}
\label{sec:experiments}

\subsection{Recognition Experiment}

For recognition, we tested on the KTH Actions dataset.  The set contains six different actions (\emph{boxing}, \emph{handclapping}, \emph{handwaving}, \emph{jogging}, \emph{running}, and \emph{walking}), each performed several times by $25$ different actors in four distinct scenarios: outdoors ($s1$), outdoors with scale variation ($s2$),  outdoors with different clothes ($s3$), and indoors ($s4$) \cite{Schuldt2004}.  There are a total of $2,391$ action sequences.

We followed the same partitioning scheme outlined by \citeauthor{Schuldt2004} \shortcite{Schuldt2004} to divide the dataset with respect to actors.  The training and validation sets each contain video sequences for $8$ unique actors.  The test set contains sequences from the $9$ remaining actors.  However, since our method does not differentiate between the training and validation phase, both the training and validation sets (16 actors) were used to generate training examples.

Results from our recognition experiment can be seen in Table~\ref{table:cmat}.  Using frequency features derived from optical flow and articulated pose features, we achieved an overall accuracy rate of $82.7\%$.  While this is not competitive with known state-of-the-art performances (see Table~\ref{table:compare}), it does show that our method works as it yields good performance rate.

\begin{table}[t!]
\begin{center}
\begin{tabular}{c|c*5{>{\centering\arraybackslash}m{\cellwidth}}|}
\noalign{\gdef\w#1{\multicolumn{1}{c}{#1}}}
\w{} & \w{bx} & \w{cl} & \w{wv} & \w{jg} & \w{rn} & \w{wk} \\ \hhline{~*6{|-}|}
box & \ccol{0.81} & \ccol{0.12} & \ccol{0.06} & 0.00 & 0.00 & \ccol{0.01} \\
clap & \ccol{0.06} & \ccol{0.91} & \ccol{0.02} & 0.00 & \ccol{0.01} & 0.00 \\
wave & \ccol{0.02} & \ccol{0.13} & \ccol{0.85} & 0.00 & 0.00 & 0.00 \\
jog & 0.00 & 0.00 & 0.00 & \ccol{0.84} & \ccol{0.08} & \ccol{0.08} \\
run & 0.00 & 0.00 & 0.00 & \ccol{0.38} & \ccol{0.62} & \ccol{0.01} \\
walk & \ccol{0.01} & 0.00 & 0.00 & \ccol{0.04} & \ccol{0.01} & \ccol{0.94} \\ \hhline{~*6{|-}|}
\noalign{\global\let\w\undefined}
\end{tabular}
\end{center}
\caption{Performance confusion matrix for our frequency forest model on the KTH Actions dataset.  The overall accuracy rate is $82.7\%$.}
\label{table:cmat}
\end{table}

\rowcolors{2}{gray!15}{white}
\begin{table}[tbh]
\begin{center}
\begin{tabular}{| l | c |} \hline 
Method & Accuracy (\%) \\ \hline \hline
Schuldt et al. [2004] & 71.7\% \\
Dollar et al. [2005] & 81.2\% \\
{\bf Frequency Forest} & {\bf 82.7\%} \\
O'hara and Draper [2012] & 97.9\% \\
Sadanand and Corso [2012] & 98.2\% \\ \hline
\end{tabular}
\end{center}
\caption{Recognition accuracies on the KTH Actions dataset for various known methods, using Schuldt's training/testing partitioning of the data.}
\label{table:compare}
\end{table}
\rowcolors{2}{}{}

Furthermore, the results show that we do well in recognizing the different hand-movement actions (e.g. \emph{boxing}, \emph{handclapping}, and \emph{handwaving}).  Our system also did a good job at recognizing \emph{walking}.

Our biggest hurdle comes from distinguishing between \emph{running} and \emph{jogging}.  Our system mistakenly labelled $38\%$ of the run videos as jogging.  This confusion can partly be explained by the fact that we were operating in the frequency domain, where the rate of body movements between running and jogging are often hard to differentiate.  This is especially true across different actors, where one person's rate of jogging may be indistinguishable from another person's rate of running.  Perhaps in cases like these, other methods might profitably augment frequency-based classification features.

\subsection{Robustness to Variability}

To show robustness, we set up an experiment that involves training and testing on different sets of data that vary in size and complexity/variability.  These sets are formed by mixing and combining data from different video scenarios in the KTH Actions dataset.

Intuitively, we expect that if the test set is held constant, then the increase in size and variability of the training set would result in more informative examples for the classifiers to train on.  This should generally produce better and more reliable action models and positively influence recognition performance.

On the other hand, if we hold the training set constant and increase the size and variability of the test set, then it should make the test set more challenging for the classifiers.  This should in turn negatively affect performance.  However, if the features used to train the models are robust to variability, then one would expect the negative impact of an increase in variability in the test set to be small.  That is to say recognition performance using these robust features should stay relatively stable and does not significantly decrease as the variability of the data in the test set increases.

For this experiment, we created different training sets by mixing data from different scenarios for all different combinations of scenarios.  For testing, we adopted the four scenario configurations used by ~\citeauthor{Schuldt2004} \shortcite{Schuldt2004}: $\{s1\}$, $\{s1,s4\}$, $\{s1,s3,s4\}$, $\{s1,s2,s3,s4\}$.  Clearly, test configurations that include more scenarios are \emph{harder} not only because of an increase in the number of test instances, but also due to the increase in variability between data from different scenarios.  Furthermore, we retain the same actor-based partitioning scheme as above, which means we never train and test on the same actor.  Thus, there is always an inherent variability in the actor involved, even for data from the same scenario.

\begin{figure}[!ht]
\centering
\includegraphics[width=1\linewidth]{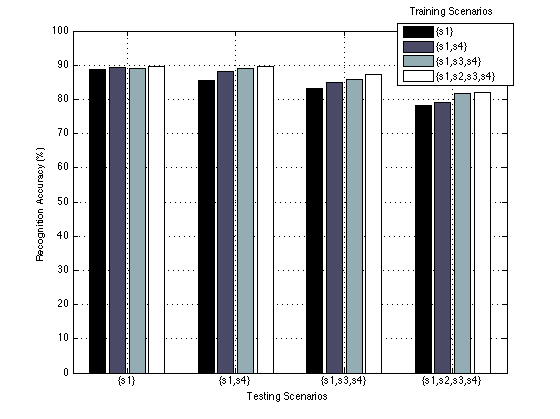}
\caption{Recognition accuracies for four representative training configurations under different test sets.  Performance for each training configuration drops slightly as the test set increases in size and complexity. Results shown are averaged across three independent runs}
\label{fig:test-varying}
\end{figure}

Figure~\ref{fig:test-varying} shows the recognition accuracy for four representative training configurations.  Complete results for all training combinations are shown in Table~\ref{table:test-varying}.  Each accuracy rate given is the average of three independent runs.


\begin{table}[tbh]
\begin{center}
\begin{tabular}{c|c c c c|} \cline{2-5}
~ & \multicolumn{4}{ c| }{Testing Scenarios} \\ \hline
\multicolumn{1}{ |c| }{Training Scenarios} & \textsf{\bf S1} & \textsf{\bf S2} & \textsf{\bf S3} & \textsf{\bf S4} \\ \hline \hline
\multicolumn{1}{ |c| }{$\{s1\}$} & $88.7$ & $85.4$ & $83.3$ & $78.2$ \\
\multicolumn{1}{ |c| }{$\{s2\}$} & $55.8$ & $61.8$ & $59.5$ & $63.8$ \\
\multicolumn{1}{ |c| }{$\{s3\}$} & $88.8$ & $85.2$ & $83.0$ & $76.8$ \\
\multicolumn{1}{ |c| }{$\{s4\}$} & $85.0$ & $86.4$ & $82.3$ & $75.8$ \\
\multicolumn{1}{ |c| }{$\{s1,s2\}$} & $85.4$ & $83.8$ & $83.4$ & $79.8$ \\
\multicolumn{1}{ |c| }{$\{s1,s3\}$} & $89.3$ & $87.5$ & $85.2$ & $78.4$ \\
\multicolumn{1}{ |c| }{$\{s1,s4\}$} & $89.3$ & $88.2$ & $85.0$ & $79.1$ \\
\multicolumn{1}{ |c| }{$\{s2,s3\}$} & $87.0$ & $86.1$ & $82.8$ & $79.6$ \\
\multicolumn{1}{ |c| }{$\{s2,s4\}$} & $81.7$ & $85.3$ & $82.6$ & $80.3$ \\
\multicolumn{1}{ |c| }{$\{s3,s4\}$} & $88.7$ & $88.1$ & $84.8$ & $78.1$ \\
\multicolumn{1}{ |c| }{$\{s1,s2,s3\}$} & $88.8$ & $86.9$ & $84.9$ & $80.6$ \\
\multicolumn{1}{ |c| }{$\{s1,s2,s4\}$} & $89.5$ & $88.6$ & $86.2$ & $80.6$ \\
\multicolumn{1}{ |c| }{$\{s1,s3,s4\}$} & $\boldsymbol{89.9}$ & $88.3$ & $86.8$ & $79.1$ \\
\multicolumn{1}{ |c| }{$\{s2,s3,s4\}$} & $89.0$ & $89.0$ & $85.8$ & $81.8$ \\
\multicolumn{1}{ |c| }{$\{s1,s2,s3,s4\}$} & $89.6$ & $\boldsymbol{89.5}$ & $\boldsymbol{87.3}$ & $\boldsymbol{82.1}$ \\ \hline
\end{tabular}
\end{center}
\caption{Recognition accuracies for different combinations of train and test sets.  For each row of training configuration, increasing the test set complexity and size does not significantly affect performance.  Note: \textsf{\bf S1} = $\{s1\}$, \textsf{\bf S2} = $\{s1,s4\}$, \textsf{\bf S3} = $\{s1,s3,s4\}$, \textsf{\bf S4} = $\{s1,s2,s3,s4\}$.  Results shown are averaged across three independent runs. Bold font indicates the best score in a given column.}
\label{table:test-varying}
\end{table}

Results shown in Figure~\ref{fig:test-varying} are representative of the general trends that we observed, and also validate our expectations of the system.  As we increase the size and variability of the test set, performance tends to drop across the board, regardless of the training configuration.  However, this decrease in performance is relatively small, which supports the robustness property of our features.  In addition, the graph also shows that increasing the training set to cover more scenarios does positively increase performance slightly.  That is to say, more training does help.

The complete set of results in Table~\ref{table:test-varying} also generally follows the described trends except for a few odd cases, most of which involve the inclusion of scenario $s2$ in training and/or testing.  This is not very surprising as scenario $s2$ is significantly different than the others.  We think that the variability introduced by the variation in scales in $s2$ is often so overwhelming that it dominates, and perhaps alters, the actual underlying signal of the action, resulting in something that looks very much different than the same action from another scenario.  As such, we believe that it would take, on average, more training data to successfully learn a good action model when using examples from $s2$.  Hence, results from small training sets containing $s2$ may be unpredictable and difficult to interpret.

Additionally, there are few other odd cases that slightly contradict the general trends due to the overlapping, or lack there of, between scenarios in the train and test sets.  For example, when training on scenario $s4$ and testing on scenario $s1$, we performed at $85.0\%$.  When we increase the test set to include another scenario, one would expect the performance to either remain stable or decrease.  However, when the added scenario is also $s4$, generating a test set that includes $\{s1,s4\}$, the performance actually increased slightly to $86.4\%$, contradicting the trend.  This, of course, can be explained by the overlapping of scenario $s4$ in both the train and test configurations.

\section{Conclusion}
\label{sec:conclusion}

This paper described a simple and efficient method to reduce the effects of data variability on recognition performance.  We demonstrated how commonly used features like optical flow and articulated pose can be transformed into frequency-domain features that are less susceptible to variability in motions, viewpoints, dynamic backgrounds, and other challenging conditions.  Furthermore, we showed how these frequency features can be used with a forest-based classifier to produce good and robust results on the KTH Actions dataset.

Although our method did not achieve the best performance in comparison with other known state-of-the-art systems (Table~\ref{table:compare}), we did achieve consistent performance that did not decrease much when variability in the test set increases.  We attribute this stability in performance of our system to the robustness of our frequency-domain features. 

We note that the presented method is not limited to the optical flow and articulated pose features that are described in the paper.  We believe this technique can used with any temporal data series, and plan to show the benefits of operating in the frequency domain for other recently developed action recognition features.  In addition, we will further explore the extent to which frequency domain features are robust to variability.  One of the first steps in this line of work will be to formally define the meaning of feature robustness and to formulate a way to accurately measure it.  We are also interested in evaluating for robustness using cross-dataset testing, similar to the experiment described by~\citeauthor{Cao2010} \shortcite{Cao2010}.  We believe that our robust frequency features should do well in recognizing the same action across many different datasets and plan to test this hypothesis in future work.

\bibliographystyle{aaai}
\bibliography{bibliography}

\end{document}